\documentclass{article}

\PassOptionsToPackage{numbers, compress}{natbib}

\usepackage[preprint]{style/neurips}

\usepackage{amsmath,amsfonts,bm}

\usepackage[utf8]{inputenc} 
\usepackage[T1]{fontenc}    
\usepackage{hyperref}       
\usepackage{url}            
\usepackage{booktabs}       
\usepackage{nicefrac}       
\usepackage{microtype}      
\usepackage[table]{xcolor}         
\usepackage{graphicx}
\usepackage{multirow}
\usepackage{subcaption}
\usepackage{pifont} 
\usepackage{cleveref}
\usepackage{adjustbox}
\usepackage{wrapfig}
\usepackage{colortbl}

\newcommand{\cmark}{\ding{51}} 
\newcommand{\xmark}{\ding{55}} 
\newcommand{\inc}[1]{\textsuperscript{\tiny\textcolor{green!60!black}{$\uparrow$#1}}}
\newcommand{\dec}[1]{\textsuperscript{\tiny\textcolor{red}{$\downarrow$#1}}}
\newcommand{\decg}[1]{\textsuperscript{\tiny\textcolor{green!60!black}{$\downarrow$#1}}}
\renewcommand{\paragraph}[1]{\noindent {\bf #1}}
\renewcommand{\paragraph}[1]{\vspace{0.2em} \noindent {\bf #1}}

\title{Scaling Up Audio-Synchronized Visual Animation: An Efficient Training Paradigm}

\author{%
  \textbf{Lin Zhang}$^1$, 
  \textbf{Zefan Cai}$^1$, 
  \textbf{Yufan Zhou}$^3$, 
  \textbf{Shentong Mo}$^2$, 
  \textbf{Jinhong Lin}$^1$, 
  \textbf{Cheng-En Wu}$^1$,\\ 
  \textbf{Yibing Wei}$^1$,
  \textbf{Yijing Zhang}$^1$, 
  \textbf{Ruiyi Zhang}$^4$, 
  \textbf{Wen Xiao}$^5$, 
  \textbf{Tong Sun}$^4$, 
  \textbf{Junjie Hu}$^1$, \\
  \textbf{Pedro Morgado}$^1$ \\
  $^1$University of Wisconsin Madison,
  $^2$Carnegie Mellon University \\
  $^3$Luma AI,
  $^4$Adobe Research,
  $^5$Microsoft
}

\begin{document}

\maketitle

\begin{abstract}


Recent advances in audio-synchronized visual animation enable control of video content using audios from specific classes.  However, existing methods rely heavily on expensive manual curation of high-quality, class-specific training videos, posing challenges to scaling up to diverse audio-video classes in the open world.
In this work, we propose an efficient two-stage training paradigm to scale up audio-synchronized visual animation using abundant but noisy  videos. In stage one, we automatically curate large-scale videos for pretraining, allowing the model to learn diverse but imperfect audio-video alignments. In stage two, we finetune the model on manually curated high-quality examples, but only at a small scale, significantly reducing the required human effort.
We further enhance synchronization by allowing each frame to access rich audio context via multi-feature conditioning and window attention.
To efficiently train the model, we leverage pretrained text-to-video generator and audio encoders, introducing only 1.9\% additional trainable parameters to learn audio-conditioning capability without compromising the generator's prior knowledge. 
For evaluation, we introduce AVSync48, a benchmark with videos from 48 classes, which is 3$\times$ more diverse than previous benchmarks. Extensive experiments show that our method significantly reduces reliance on manual curation by over 10$\times$, while generalizing to many open classes.

\end{abstract}
\section{Introduction}
\label{sec:intro}

\begin{figure}[t]
    \centering
    \includegraphics[width=\textwidth]{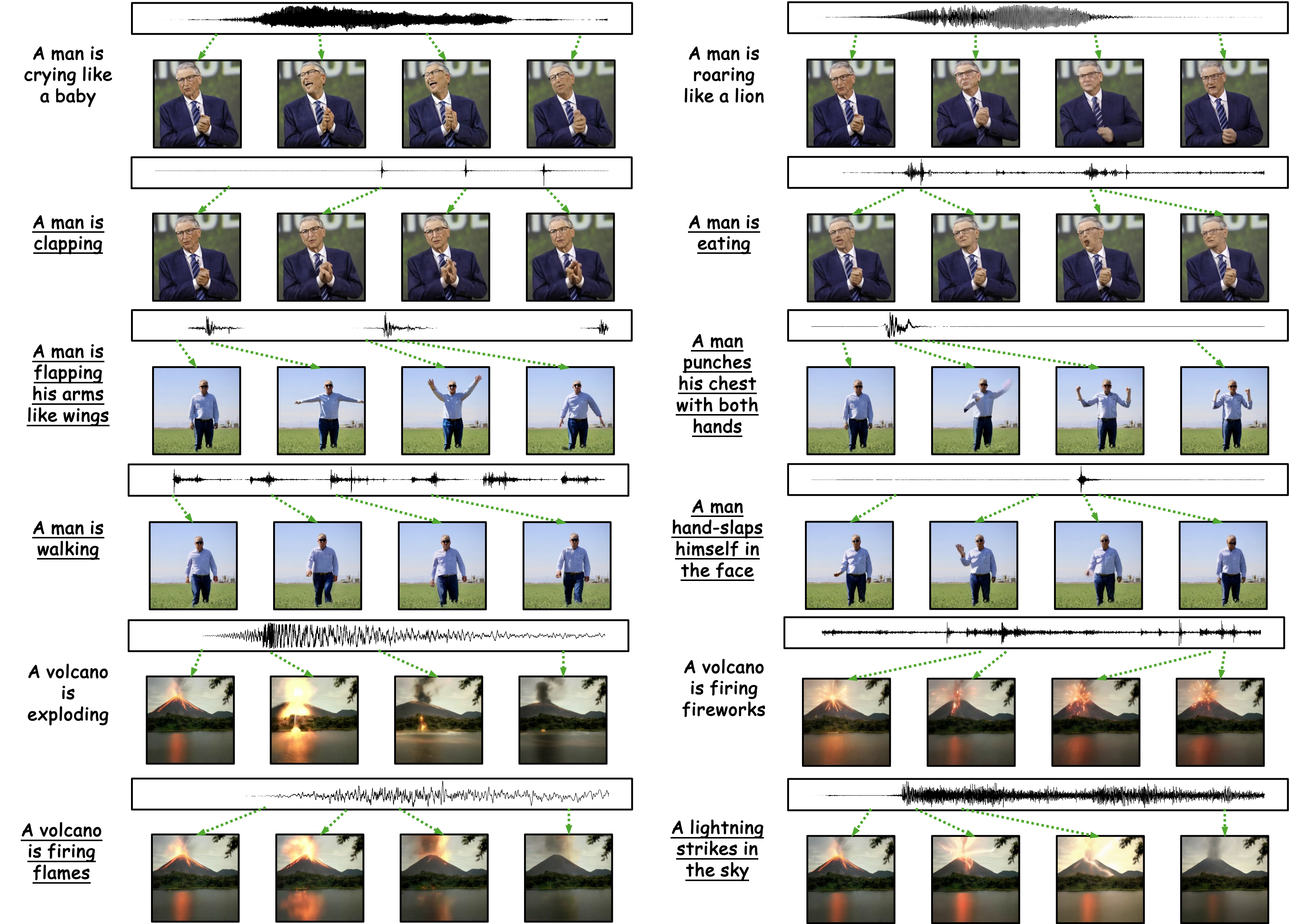}
    \caption{Generation results on open-domain images, captions, and audios from internet. Out-of-domain controls are underlined. Results explained in \Cref{sec:ablation open domain}. More videos provided in Appendix.}
    \label{fig:open-domain}
\end{figure}

Text-to-Video generation (T2V)~\cite{xing2023dynamicrafter,wang2023modelscope,vid2vid-zero,singer2022makeavideo,chen2023videocrafter1,text2video-zero,polyak2025moviegen,HaCohen2024LTXVideo,kong2024hunyuanvideo,wan2025} has progressed rapidly, driven by large-scale diffusion models~\cite{ho2020ddpm,song2022ddim,robin2022ldm,crowson2024sd3} trained on massive text-video datasets~\cite{Bain21webvid10m,xue2022hdvila,chen2024panda70m,wang2024internvidlargescalevideotextdataset}. These models generate smooth, visually compelling videos with strong semantic alignment to captions.
In contrast, Audio-Synchronized Visual Animation (ASVA)~\cite{linz2024asva,wang2025keyvid,yariv2023tempotoken}, which aims to animate images based on diverse audio cues, remains underexplored and is largely confined to narrow domains like talking face generation~\cite{lipreading,zhou2019talking,park2022synctalkface,outoftime,ng2024audio2photoreal,ye2023geneface,Zhou2021Pose}. This limited scope overlooks ASVA’s potential across broader domains across physical actions, sound effects, and musical performances, etc.

In this paper, we ask a question:  
\textbf{How to scale up ASVA to diverse audio-visual domains in an efficient and generalizable manner?}  
We approach this question by addressing two core challenges.


\textbf{Challenge 1: What is a scalable training paradigm for ASVA?}
Scaling Audio-Synchronized Visual Animation (ASVA) to open-domain scenarios requires training data that satisfies both quantity and quality demands. While quantity refers to the breadth and volume of data, quality pertains primarily to the precise synchronization between audio and the corresponding visual actions—typically governed by physical realism~\cite{chen2021vggsoundsync,ebenezer2021detectionavsync,sparse2022iashin,Chung2019perfectmatch,NEURIPS2018coopav,aytar2016soundnet}.
On the quantity front, large-scale and diverse video datasets covering a wide range of audio-visual patterns are readily available online~\cite{chen2020vggsound,audiocaps,audioset}, making this requirement relatively easy to satisfy. However, meeting the quality criterion is significantly more challenging. As illustrated in \Cref{fig:paradigm}, visually, the video should be free from issues like low resolution, excessive text overlays, shot transitions, or abrupt camera movements; acoustically, the audio should be clean and clearly aligned with visual events at the frame level, without interference from ambient noise, occluded sound sources, unrelated visual motions, or other misalignments.
While certain types of visual noise can be mitigated using automated filtering tools as in text-to-video pipelines~\cite{chen2024panda70m,kong2024hunyuanvideo}, audio-visual misalignments are often more subtle and typically require costly manual curation~\cite{linz2024asva}. This creates a fundamental dilemma: increasing data volume enhances diversity and coverage, but maintaining high-quality audio-visual synchronization becomes increasingly labor-intensive at scale.

Talking-face generation~\cite{lipreading,zhou2019talking,park2022synctalkface} sidesteps this issue by focusing on a limited domain with abundant, naturally aligned human-centric content (e.g., interviews, podcasts) and clean audio-action synchronization cues. However, as we expand into broader domains, the complexity of audio-visual synchronization patterns grows significantly. To address this challenge, prior work~\cite{linz2024asva} extended ASVA to 15 classes, trading off quantity for synchronization quality via extensive manual curation, which cannot scale. A new training paradigm is therefore needed to scale up ASVA efficiently.




Our first contribution introduces a two-stage training paradigm addressing both quantity and quality requirements by combining 1) large-scale pretraining with 2) few-shot fine-tuning, where the two stages operate on opposite ends of the quantity-quality spectrum.
In the first stage, we deploy off-the-shelf models and pre-processing techniques to automatically remove visual noise from a large scale dataset, which is then used for pretraining. Although pretraining data still contains many imperfect audio-video alignments, this allows the model to learn a wide range of synchronization patterns across diverse classes. 
Then, in the second stage, we fine-tune the model on a small set of manually curated clean examples to enhance generation quality and synchronization. This two-stage paradigm significantly reducing the budget for exhaustive manual data curation, thus allowing efficient generalization to diverse audio and video classes.

\textbf{Challenge-2: How to train an ASVA model efficiently?}
ASVA and T2V share key objectives: producing visually appealing, coherent, and temporally smooth videos. However, ASVA introduces the additional challenge of aligning the generated motion with fine-grained audio cues. Given the availability of powerful T2V models, a natural question arises: can the models be efficiently extended with audio control while preserving strong video generation capabilities?

Our second contribution addresses this question through lightweight and efficient audio conditioning modules on top of existing T2V models. To capture complementary audio information at different granularities, we extract a mix of features across multiple pretrained encoders and at various layers, resulting in a sequence of temporal audio tokens.
We then introduce audio cross-attention modules into the pretrained generation model with window attention~\cite{vaswani2023attention}. Instead of relying on a rigid many-to-one conditioning scheme between audio tokens and frames as in current methods, our window attention allow each frame to attend to a broader temporal audio context, with some audio tokens shared between frames. 
These conditioning modules adds minimal trainable parameters for enabling synchronized audio-driven control, while preserving the visual strengths of the base T2V model.

To rigorously evaluate the proposed training strategy and conditioning mechanism, our third contribution is AVSync48: a new benchmark comprising 48 diverse audio-visual classes. Each class includes 30 high-quality videos with clean and well-aligned audio-motion synchronization. AVSync48 triples the number of categories found in existing benchmarks~\cite{linz2024asva,owens2016greatesthits}, enabling evaluation across a significantly broader range of audio-visual domains.

With the three contributions,  we present a scalable, efficient, and high-performing framework for audio-synchronized visual animation, bridging the gap between existing models in extremely restricted domains and an open-domain model. Experiments show our model can significantly reduce dependence on manual data curation by over 10$\times$ on both AVSync15~\cite{linz2024asva} and AVSync48. Our model surpasses state-of-the-arts~\cite{tang2023codi,yariv2023tempotoken,linz2024asva,wang2025keyvid,lee2023aadiff,jeong2023tpos} by a significant margin on both benchmarks, while being much more efficient with only $1.9\%$ trainable parameters. More importantly, we achieve strong generalization on many open audio-visual classes, allowing for accurate image animation controlled by both text and audio, as shown in \Cref{fig:open-domain}.


\section{Related work}
\label{sec:related-work}

\subsection{Audio-to-video generation}
Many works on Audio-to-Video generation (A2V)~\cite{aist-dance-db,li2021aist++,jeong2023tpos,tang2023codi,lee2022soundguided} attempted to generate videos with only semantics aligned with input audio, particularly in the music and environment sounds domain.
However, these works often overlook the importance of temporal alignment, also known as synchronization, where audio and video should adhere to physical timing constraints.
Conventionally, due to the huge commercial potential and availability of abundant clean training videos, tremendous synchronized A2V works focused on the single fine-grained  human talking~\cite{lipreading,zhou2019talking,park2022synctalkface,outoftime,ng2024audio2photoreal,ye2023geneface,Zhou2021Pose} domain. To expand to diverse domains beyond human talking, Audio-Synchronized Visual Animation (ASVA)~\cite{linz2024asva,wang2025keyvid} was recently proposed. The initial work on ASVA~\cite{linz2024asva} identified the major challenge to be the scarcity of clean videos  demonstrating clear audio-motion synchronization cues for training, and introduced a manually curated dataset AVSync15. Later works~\cite{wang2025keyvid} attempted to improve benchmark performance via two-stage generation, i.e.,  adaptive key-frame generation followed by dense frame interpolation. 
In this work, by adopting our proposed efficient training strategies, we develop a model that not only significantly improves benchmark performance but also generalizes effectively across many open classes.


\subsection{Budget-efficient data collection}

 There has been extensive discussion on data collection a under limited human efforts(or budget), with focus on balancing the low-quality but abundant uncurated data and high-quality but expensive manually collected data.
 Many paradigms have been proposed to solve this quantity vs.~quality issue, such as unsupervised~\cite{chen2020simpleframeworkcontrastivelearning,he2019moco,chen2020mocov2,he2021maskedautoencodersscalablevision,xie2022simmimsimpleframeworkmasked}, semi-supervised~\cite{tarvainen2018meanteachersbetterrole,laine2017temporalensemblingsemisupervisedlearning,sohn2020fixmatchsimplifyingsemisupervisedlearning,berthelot2019mixmatchholisticapproachsemisupervised,NIPS2004entropymin}, few-shot~\cite{Koch2015SiameseNN,vinyals2017matchingnetworksshotlearning,sung2018learningcomparerelationnetwork,snell2017prototypicalnetworksfewshotlearning}, and active learning~\cite{active1,active2,sener2018activelearningconvolutionalneural}.  In the context of ASVA, such a quantity vs.~quality issue becomes more severe due to the overwhelming noise of sounded videos. Previous works~\cite{linz2024asva,wang2025keyvid} discarded large-scale noisy videos and simply relied on expensive, high-quality data for training on a small domain, which restricts scalability to a wider range of audio-video classes in reality. In this work, we demonstrate that pretraining on large-scale curated noisy videos can largely reduce high-quality data needed. Our approach enables scaling to a much wider variety of audio and video classes within the same budget, advancing the development of an open-domain ASVA model.

\subsection{Efficient adaptation of pretrained models}

Training models from scratch is expensive. Instead, many existing works attempted to adapt large pretrained models~\cite{vicuna2023,robin2022ldm} for new tasks by utilizing their prior knowledge. 
One approach focuses on learning task-specific knowledge within a particular domain by allowing for finetuning to the original parameters of the pretrained model~\cite{hu2021loralowrankadaptationlarge,liu2023llava,zhu2023minigpt4enhancingvisionlanguageunderstanding,zhao2023motiondirector,wu2023lamp,linz2024asva,wang2025keyvid,tan2024ominicontrol,tan2025ominicontrol2}, which however often compromises the pretrained model's prior knowledge. 
The other approach is to completely freeze the pretrained model, and augment it with additional trainable layers~\cite{li2023gligen,zhang2023adding,mou2023t2iadapterlearningadaptersdig}. While this preserves the pretrained knowledge, it requires careful design to avoid excessive trainable parameters, which can hinder training efficiency. 
In this work, we adopt the latter approach and find that only a small number of trainable parameters are sufficient to achieve high performance. Specifically, we build upon a fully frozen text-guided image animation model~\cite{xing2023dynamicrafter} and audio encoders~\cite{girdhar2023imagebind,Chen2022beats}, and introduce additional lightweight projection layer and audio cross attention layers for training, totaling only 55M parameters(or 1.9\% of the whole model). Leveraging the strong priors in the pretrained models, our method not only achieves state-of-the-art performance on benchmarksm but also generalizes effectively to a man open classes.
\section{Two-stage learning paradigm}
\label{sec:learn-paradigm}

\begin{figure}[!thp]
    \centering
    \includegraphics[width=0.75\textwidth]{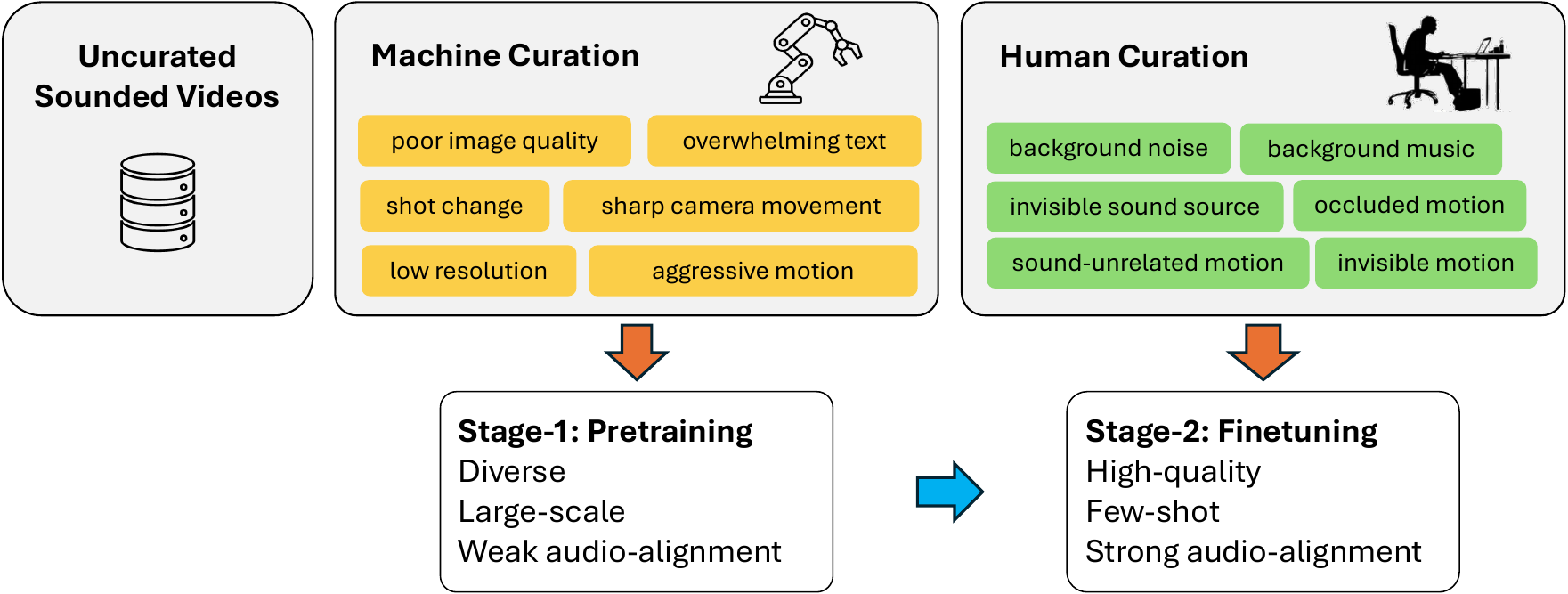}
    \caption{Data collection and training paradigm. We automatically curate video datasets to obtain diverse and large-scale pretraining data, which is however still noisy in ambiguous synchronization patterns. We then allow minimal human efforts to collect a few high-quality data points for finetuning.}
    \label{fig:paradigm}
\end{figure}

Scaling up Audio-Synchronized Visual Animation (ASVA) requires collecting videos with consideration of three key factors: (1) Quantity: the data should be acquired at scale, covering a diverse and open set of categories; (2) Quality: videos should exhibit clear temporal synchronization between audio and visual motions to provide effective frame-level supervision; (3) Budget: data collection should be cost-effective and minimally labor-intensive. Prior work~\cite{linz2024asva,wang2025keyvid} achieved strong performance on downstream ASVA benchmarks by hinging on manually curated datasets, such as 90 training video for a single class on AVSync15. Given the virtually infinite variety of audio-visual events~\cite{audioset,chen2020vggsound}, and the high budget of human labeling, this approach does not scale. 
Instead, we propose a more scalable alternative: leveraging large volumes of automatically curated videos for pretraining, which trades off audio-visual alignment quality for scale and diversity, followed by fine-tuning on a small number of high-quality, manually verified samples, as shown in \Cref{fig:paradigm}.

\paragraph{Data curation for ASVA pretraining}
Instead of directly training on uncurated noisy videos, we apply automatic filters to curate the training data distribution without affecting diversity and scale. Specifically, we (1) remove videos that are too short or have low FPS or resolution; (2) use PySceneDetect~\cite{pyscenedetect} to break videos into semantics-consistent scenes; (3) use optical flow detection~\cite{teed2020raftrecurrentallpairsfield} to discard video segments with overwhelmingly sharp motions; (4) apply text detection~\cite{craft2019} to discard videos with overwhelming onscreen text. In addition to these steps, we also tried other filters~\cite{llama3modelcard,liu2023llava,Liu_2024_CVPR_llava15,mei2023wavcaps,laionaesthetics}, but found they can reduce dataset diversity significantly, and thus decided not to deploy them in the final curation pipeline. More details on automatic curation can be seen in the appendix.
In practice, we deployed the above curation steps on VGGSound~\cite{chen2020vggsound}(170K videos), while also ensuring no overlap with AVSync15 and AVSync48 test sets. This process resulted in a pretraining dataset containing 152K videos. Although noisy audio-visual alignment still exists, the model can be supervised to learn audio-driven visual animation across wider and more diverse data domains.

\paragraph{Budget-efficient finetuning on high-quality data}
The pretraining data distribution covers a wide range of audio-visual classes, but high-quality samples with strong temporal alignment are not prominent in the dataset. As a result, the pretrained model is suboptimal regarding precise audio control of video motions.
Therefore, to enhance the precision of audio control, we further collect a small set of high-quality videos for finetuning. With pretraining, the number of high-quality videos required, as well as the budget used for manually collecting the videos, can be largely reduced.

\paragraph{AVSync48 benchmark.}
To validate our training paradigm, we manually curated a benchmark consisting of 48 categories of diverse audio-video motions derived from VGGSound~\cite{chen2020vggsound} and AudioSet~\cite{audioset}. First, we merged similar classes from both datasets. Next, we manually reviewed several videos per class, eliminating classes with are noisy in audio-visual alignment by nature. We then recruited annotators to manually identify high-quality clips from each class according to stringent criteria. To expedite this process, we developed a QuickTime Player~\cite{quicktime} plugin that enabled annotators to record valid video segments using keyboard shortcuts. Each recorded clip was further validated by another annotator for quality assurance. In total, we manually identified 30 high-quality videos for each of the 48 audio events.
The collected benchmark, denoted AVSync48, is intended to assess the effectiveness of AVSA models on a broader set of classes than currently available benchmarks. To support evaluations through few-shot domain-specific fine-tuning, we further provide an official train/test split, comprised of 10/20 videos per class for training/testing, respectively.

We experimentally verified that, compared to previous efforts that only rely on manually curated training data, our paradigm significantly reduced the amount of high-quality samples in need by a factor of over 10 (see~\Cref{tab:fsl}), while being generalizable to open domains (see~\Cref{fig:open-domain}). Overall, with reduced human annotation requirements but still competitive performance, our two-stage training paradigm can learn a broader range of classes, paving the way for more open-domain generation.
\section{Model architecture}
\label{sec:arch}

\begin{figure}[!t]
    \centering
    \begin{subfigure}[b]{0.53\textwidth}
        \includegraphics[width=\textwidth]{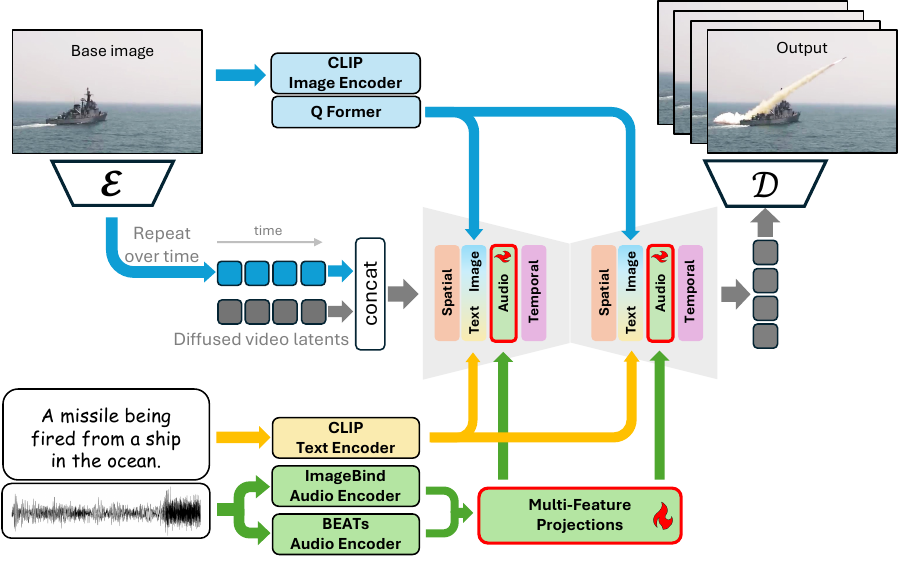}
        \caption{Architecture overview.}
        \label{fig:arch}
    \end{subfigure}
    \hfill
    \begin{subfigure}[b]{0.42\textwidth}
        \includegraphics[width=\textwidth]{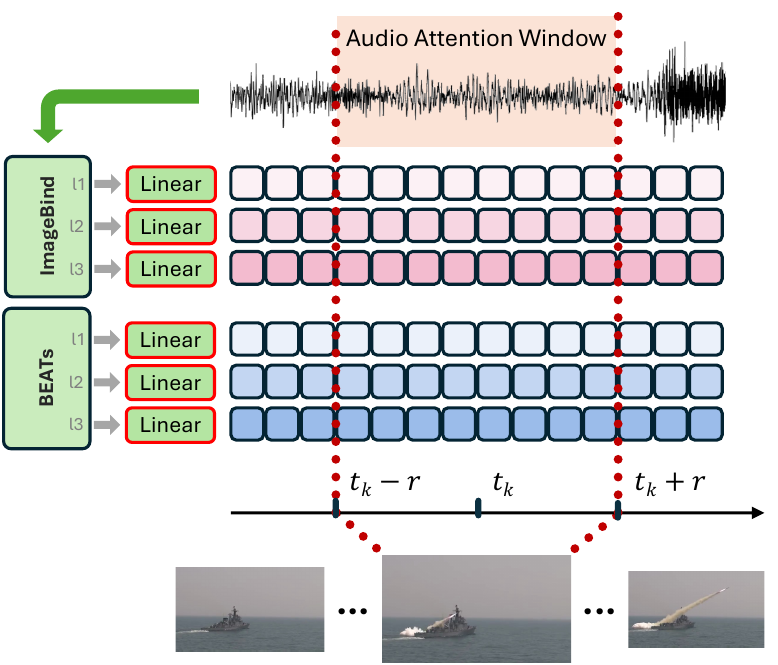}
        \caption{Audio conditioning modules.}
        \label{fig:modules}
    \end{subfigure}
    \caption{(a) Architecture overview. We build our method on a frozen pretrained text guided image animation model, DynamiCrafter, and add additional audio cross-attention layers right before temporal attention layers. We only train minimum parameters ($1.8\%$), including the multi-feature projection layers and audio cross-attention layers, as marked in \textcolor{red}{red}. (b) multi-feature and window attention for audio conditioning. We collect audio features in different granularities from frozen ImageBind and BEATs encoders, providing comprehensive audio signals for conditioning. At the frame level, each frame is conditioned on all audio features within its local window of radius r, allowing for larger audio context overlapping with its nearby frames.}
    \label{fig:overview}
\end{figure}

ASVA requires not only generating visually coherent and physically plausible video sequences, as in text-to-video models, but also achieving fine-grained synchronization between visual motion and the accompanying audio. This dual objective poses significant challenges in open-domain settings, where the model must generalize across diverse and unconstrained content. To address this, we build our system on top of pretrained video generation models offering strong priors over video dynamics, and audio encoders that are effective for in-the-wild audio understanding. To retain their prior knowledge as well as training efficiency, we keep these pretrained models completely frozen, and focus on augmenting them with strong capabilities to understand and generate synchronization patterns between audio and video frames via lightweight and trainable audio conditioning modules. In the following sections, we detail how these pretrained models are integrated and extended to support ASVA, as shown in \Cref{fig:overview}.

\paragraph{Frozen T2V base model} We built our model upon \textbf{DynamiCrafter}~\cite{xing2023dynamicrafter}, a latent video diffusion model for text-guided visual animation task. It comprises of a VAE encoder and decoder, a denoising UNet, a CLIP text encoder, a CLIP image encoder, and a Q-Former projector. The input image is conditioned in two ways: (1) it is encoded by the CLIP image encoder followed by Q-Former into features for cross attention; (2) it is encoded by the VAE encoder into latent, then repeated along the time axis to be concatenated with noisy videos latents in feature dimension. The model was trained on large scale text-video pairs~\cite{Bain21webvid10m}, providing strong video generation priors.

\paragraph{Audio multi-feature conditioning}
Prior works~\cite{linz2024asva,wang2025keyvid} extracted audio features from a single layer in a pretrained audio encoder. However, different audio encoders capture distinct aspects of audio semantics and structures, depending on their training objectives. To exploit their complementary strengths, we adopted two audio encoders:
(1) \textbf{ImageBind}~\cite{girdhar2023imagebind} a cross-modal encoder contrastively trained to align audio features with CLIP's\cite{radford2021CLIP} image representations. This offers strong semantic grounding, enabling high-level audio events to influence video generation in a conceptually aligned way.
(2) \textbf{BEATs}~\cite{Chen2022beats} a self-supervised encoder trained with masked audio modeling~\cite{huang2022amae,MaskedAutoencoders2021} and iterative clustering. Trained to predict masked audio tokens, BEATs' representations potentially provide more fine-grained temporal cues and low-level rhythmic patterns, which are crucial for synchronization patterns. 
We further enrich these representations by extracting multi-scale features from low, mid, and high layers in both encoders, capturing a wider range of audio characteristics. This multi-encoder, multi-scale design provides a diverse and robust set of audio features, enhancing the model’s ability to learn fine-grained audio-visual synchronization cues. In practice, we introduce a linear layer after each extracted feature to project it onto latent space for cross attention.

\paragraph{Audio window attention} 
ASVA requires synchronization at the frame level. To do so, we allow each frame to attend to its local audio tokens via introduced audio cross attention layers. Consider a sequence of $K$ generated frames at timestamps $\{t_1, t_2, \ldots, t_K\}$, where consecutive frames are separated by a time gap of $1/\text{\textit{FPS}}$ seconds. Each frame should be aware of sufficient audio context around itself to learn accurate synchronization. One important design question here is the context size. Broader contexts might result in less specificity to synchronization cues with too much overlapping between consecutive frames' vision, while smaller contexts might result in incomplete coverage of conditioning signals. Prior works heuristically forced a many-to-one mapping between audio tokens and frames, where the audio tokens are uniformly split into non-overlapping $K$ chunks, thus allowing the $i$-th frame to attend to audio tokens within its local time window of radius $r=0.5/\text{\textit{FPS}}$ seconds. We instead allow for broader audio context window, i.e., larger $r$. The aforementioned specificity-coverage trade-off was verified experimentally, with the optimal radius in our experiments being  $r=1.5/\text{\textit{FPS}}$, i.e., each frame shares some audio tokens with its nearby frames.

We insert the audio cross attentions before each temporal attention blocks to provide more frame-level audio supervision before different frames interacting with each other. We apply classifier-free guidance for training, randomly dropping image, text, and audio tokens with $5\%$ chance. During inference, we apply multi-modal guidance scales as in DynamiCrafter. In practice, we freeze all pretrained models including DynamiCrafter, ImageBind, and BEATs, and only train the newly introduced multi-feature projection layers and audio cross attention layers, comprising of only 55M parameters, or $3.8\%$ of the denoising UNet (1.4B) and $1.9\%$ of the overall model (2.8B). In experiments, we trained two models: a 6-FPS model generating 12 frames as in previous works~\cite{linz2024asva}, and a 12-FPS model generating 24 frames for higher video quality. 
Despite the minimal trainable parameters, our model outperforms prior works trained with more parameters (see \Cref{tab:benchmark}), while demonstrating strong generalization to many open classes, thanks to the frozen priors.

\section{Experiments}%
\label{sec:experiments}

We evaluate our model across standard ASVA benchmarks. Experiments include comparisons with prior works, several ablations, and tests under few-shot and open-domain settings. Additional implementation details and \textbf{numerous generated video files} are provided in the Appendix.

\subsection{Implementation details}
\label{subsec:implementations}

\paragraph{Training and inference} Our model builds on pretrained DynamiCrafter~\cite{xing2023dynamicrafter}, ImageBind~\cite{girdhar2023imagebind}, and the BEATs audio encoder~\cite{Chen2022beats}. During all training, we uniformly sample 2-second video clips. Audio is sampled at 16 kHz and converted into 128-dimensional mel-spectrograms using 25-ms windows with 10-ms shifts. Video frames are resized to 256$\times$256 and temporally resampled at 6 or 12 FPS. For pretraining, we use the Adam optimizer~\cite{kingma2017adamoptimizer} for 30 epochs with a batch size of 32 and a learning rate of 1e-4. Stage-2's finetuning runs for 500 epochs with a batch size of 16 and a learning rate of 5e-5. Training is conducted on 8 A100 GPUs for our 12-FPS model and 4 A100 GPUs for our 6-FPS model. During inference, we apply DDIM~\cite{song2022ddim} sampling with 20 steps and classifier-free guidance scales of 2.0, 2.0, 4.0 for image, text, and audio guidance, respectively. 

\paragraph{Datasets} We use the proposed auto-curated VGGSound~\cite{chen2020vggsound} dataset for pretraining. For evaluation, we leveraged two benchmark datasets covering a diverse set of classes with strong audio-visual alignment\footnote{Datasets with a single class (impact sound)~\cite{owens2016greatesthits} or only ambient sounds~\cite{lee2022soundguided} were discarded as they do not provide a challenging enough domain for ASVA evaluation.}: (1) \textbf{AVSync15}~\cite{linz2024asva}, consisting of 15 classes covering human actions, tool usage, music instruments, and animal sounds, with 90 training videos and 10 testing videos per class. (2) \textbf{AVSync48} our curated dataset with 48 classes introduced in \Cref{sec:learn-paradigm}. 

\paragraph{Evaluation metrics}
Following conventions~\cite{linz2024asva,wang2025keyvid}, we uniformly sample 3 6-FPS clips from each video for evaluation. Videos in higher FPS are downsampled to 6-FPS. Metrics include: (1) \textbf{FVD}~\cite{fvd} measuring distributional distance between generated and ground-truth video features; (2) \textbf{IA} CLIP~\cite{radford2021CLIP,girdhar2023imagebind} semantic similarity between input audio and generated frames; (3) \textbf{IT} CLIP similarity between input text captions and generated frames; (4) \textbf{RelSync}~\cite{linz2024asva}, a synchronization score comparing audio synchronization of generated animations with ground-truth; and (5) \textbf{AlignSync}~\cite{linz2024asva}, a score jointly measuring semantic alignment and synchronization. As originally defined, RelSync and AlignSync have upper bounds of 50 and 25, respectively, representing scores of ground-truth. For clarity, we rescale both metrics to the more natural 0-100 range (RelSync $\times$2, AlignSync $\times$4).

\paragraph{Baselines}
Here we highlight two strong prior works for comparison, with others listed in the Appendix. (1) \textbf{AVSyncD}~\cite{linz2024asva} built on pretrained image StableDiffusion~\cite{robin2022ldm} and ImageBind~\cite{girdhar2023imagebind}. AVSyncD introduces several learnable layers (audio cross attention, temporal convolution, temporal attention) totaling 310M trainable parameters. We report its performance on AVSync48 by training with their official code. (2) \textbf{KeyVID}~\cite{wang2025keyvid} a two-stage model based on DynamiCrafter. It first generates non-uniform keyframes from audio, followed by a separate model that interpolates 24-FPS motion. During training, it samples 24-FPS video frames as supervision to train the audio cross-attention layers, image projection layers, image cross-attention layers, and temporal attention layers, with significantly more trainable parameters. Due to incomplete publicly available code at the time of writing, we compare to KeyVID only on AVSync15 using their reported results.

\begin{table}[t]
\centering
\caption{Comparison with previous works on AVSync15 and AVSync48.}%
\label{tab:benchmark}
\begin{adjustbox}{scale=0.75}
\begin{tabular}{c l c c c c c c}
\toprule
Dataset & Model & FPS & FVD ↓ & IA ↑ & IT ↑ & RelSync ↑ & AlignSync ↑ \\
\midrule
AVSync15
& CoDi         & 4  & 1522.6 & 28.15 & 23.42 & 83.02  & 78.16 \\
& TPoS         & 6  & 1227.8 & 38.36 & \textbf{30.73} & 79.24  & 78.68 \\
& AADiff       & 6  & 978.0  & 34.23 & 28.97 & 90.96  & 88.44 \\
& AVSyncD      & 6  & 349.1  & 38.53 & 30.45 & 91.04  & 90.48 \\
& KeyVID       & 24 & 262.3  & 39.23 & 30.30 & 96.66  & 96.32 \\
\rowcolor{cyan!8}\cellcolor{white} 
& Ours         & 6  & 258.1 & \textbf{39.74} & 30.32 & 94.66  & 94.48 \\
\rowcolor{cyan!8}\cellcolor{white} 
& Ours         & 12 & \textbf{250.4} & 39.72 & 30.37 & \textbf{99.18} & \textbf{99.00}\\
\midrule
AVSync48
& AVSyncD        & 6  & 217.9 & 36.22 & 29.71 & 90.38  & 89.40 \\
\rowcolor{cyan!8}\cellcolor{white} 
& Ours           & 6  & 99.6  & \textbf{37.80}  & \textbf{31.71} & 95.04  & 94.76 \\
\rowcolor{cyan!8}\cellcolor{white} 
& Ours           & 12 & \textbf{98.1} & 37.77 & 31.69 & \textbf{98.64} & \textbf{98.32} \\
\bottomrule
\end{tabular}
\end{adjustbox}
\end{table}

\subsection{Comparisons to prior work}
We report results on AVSync15 and AVSync48 in \Cref{tab:benchmark}. At 6-FPS, our model surpasses previous methods significantly on all metrics on both AVSync15 and AVSync48. By lifting FPS to 12, our model receives denser synchronization signals, boosting RelSync and AlignSync significantly by about 4.5 on AVSync15 and about 3.5 on AVSync48. Notably, on AVSync15, despite that KeyVID was trained with even denser 24-FPS video frames and significantly more trainable layers, our 12-FPS model can still outperform it by a large margin especially on the synchronization metrics (RelSync 96.66→99.18, AlignSync 96.32→99.00), approaching ground-truth performance.


\subsection{Ablation studies}

We conduct all ablation studies on the 6-FPS model, unless otherwise noted.

\paragraph{Data curation}
\Cref{tab:curation} shows the impact of stage-1's auto-curation and stage-2's finetuning. All results use the baseline model without optimized multi-feature conditioning or window attention.
Automatic curation in the pretrainig stage yields consistent gains across metrics over pretraining with uncurated data. For example, on AVSync15, FVD improves from 381.5 to 350.9 and AlignSync from 90.96 to 91.52. Further finetuning on manually curated data brings significant improvements, including on FVD (350.9 → 261.0) and AlignSync (91.52 → 93.64). The benefits of data curation for pretraining are even more pronounced on AVSync48, which contains much less manually curated data per class for finetuning than AVSync15 (10 vs 90), representing a more practical scenario with minimal human efforts to scale up.

\begin{table}[t]
\centering
\caption{Ablation on auto-curated data for pretraining and manually curated data for finetuning. }
\label{tab:curation}
\begin{adjustbox}{scale=0.75}
\begin{tabular}{c c c l l l l l}
\toprule
Benchmark & Curation  & Finetune & FVD ↓ & IA ↑ & IT ↑ & RelSync ↑ & AlignSync ↑ \\
\midrule
\multirow{4}{*}{AVSync15}
& \multirow{2}{*}{\xmark}  & \xmark & 381.5  & 38.63  & 29.61  & 91.62  & 90.96 \\
&                          & \cmark & 263.8\decg{117.7} & 39.76\inc{1.13} & 30.31\inc{0.70} & 93.76\inc{2.14} & 93.64\inc{2.68} \\
\cmidrule{2-8}
& \multirow{2}{*}{\cmark}  & \xmark & 350.9  & 39.06  & 29.92  & 92.00  & 91.52 \\
&                          & \cmark & 261.0\decg{89.9} & 39.77\inc{0.71} & 30.33\inc{0.41} & 93.8\inc{1.80}  & 93.64\inc{2.12} \\
\midrule
\multirow{4}{*}{AVSync48}
& \multirow{2}{*}{\xmark}  & \xmark & 157.5  & 37.13  & 31.15  & 92.74  & 92.16 \\
&                          & \cmark & 107.9\decg{49.6} & 37.90\inc{0.77} & 31.67\inc{0.52} & 94.14\inc{1.40} & 93.92\inc{1.76} \\
\cmidrule{2-8}
& \multirow{2}{*}{\cmark}  & \xmark & 133.7  & 37.69  & 31.51  & 93.58  & 93.24 \\
&                          & \cmark & 104.1\decg{29.6} & 37.89\inc{0.20} & 31.66\inc{0.15} & 94.58\inc{1.00} & 94.32\inc{1.08} \\
\bottomrule
\end{tabular}
\end{adjustbox}
\end{table}

\begin{table}[t]
\centering
\caption{Few-shot finetuning results. We use various number (K) of video per class for finetuning. Superscripts shows impact of the pretraining stage. Superscripts denote performance \textcolor{green!60!black}{gains} or \textcolor{red}{losses}.}
\label{tab:fsl}
\begin{adjustbox}{scale=0.75}
\begin{tabular}{c c c l l l l l}
\toprule
Benchmark & K & Pretrain & FVD ↓ & IA ↑ & IT ↑ & RelSync ↑ & AlignSync ↑ \\
\midrule
\multirow{9}{*}{AVSync15}
& 0     
& \xmark & 338.4 & 39.06 & 29.95 & 93.08 & 92.6 \\
\cmidrule{2-8}
& \multirow{2}{*}{1}
& \xmark & 391.8 & 39.38 & 30.29 & 90.32 & 90.00 \\
&& \cmark 
& 319.7\decg{72.1} 
& 39.56\inc{0.18} 
& 30.13\dec{0.16} 
& 94.40\inc{4.08} 
& 94.28\inc{4.28} \\
\cmidrule{2-8}
& \multirow{2}{*}{5}
& \xmark & 349.6 & 39.43 & 30.27 & 92.16 & 91.84 \\
&& \cmark 
& 306.1\decg{43.5} 
& 39.63\inc{0.20} 
& 30.30\inc{0.03} 
& 94.54\inc{2.38} 
& 94.52\inc{2.68} \\
\cmidrule{2-8}
& \multirow{2}{*}{10}
& \xmark & 352.6 & 39.75 & 30.35 & 93.5 & 93.32 \\
&& \cmark 
& 292.1\decg{60.5} 
& 39.66\dec{0.09} 
& 30.22\dec{0.13} 
& 94.62\inc{1.12} 
& 94.36\inc{1.04} \\
\cmidrule{2-8}
& \multirow{2}{*}{90}
& \xmark & 277.5 & 39.76 & 30.38 & 94.42 & 94.24 \\
&& \cmark 
& 258.1\decg{19.4} 
& 39.74\dec{0.02} 
& 30.32\dec{0.06} 
& 94.66\inc{0.24} 
& 94.48\inc{0.24} \\
\midrule
\multirow{7}{*}{AVSync48}
& 0     
& \xmark & 146.3& 37.61 & 31.54 & 94.16 & 93.76 \\
\cmidrule{2-8}
& \multirow{2}{*}{1}
& \xmark & 142.2 & 37.27 & 31.57 & 92.90 & 92.36 \\
&& \cmark 
& 102.5\decg{39.7} 
& 37.73\inc{0.46} 
& 31.73\inc{0.16} 
& 94.80\inc{1.90} 
& 94.72\inc{2.36} \\
\cmidrule{2-8}
& \multirow{2}{*}{5}
& \xmark & 116.1 & 37.65 & 31.65 & 94.38 & 94.00 \\
&& \cmark 
&  97.5\decg{18.6} 
& 37.86\inc{0.21} 
& 31.73\inc{0.08} 
& 95.00\inc{0.62} 
& 94.76\inc{0.76} \\
\cmidrule{2-8}
& \multirow{2}{*}{10}
& \xmark & 116.8 & 37.71 & 31.75 & 94.48 & 94.16 \\
&& \cmark 
&  99.6\decg{17.2} 
& 37.80\inc{0.09} 
& 31.71\dec{0.04} 
& 95.04\inc{0.56} 
& 94.76\inc{0.60} \\
\bottomrule
\end{tabular}
\end{adjustbox}
\end{table}

\paragraph{Efficiency of proposed paradigm}
To validate our paradigm as scalable, efficient, and requiring less human curation efforts, we provide pretraining and K-shot finetuning results on AVSync15 and AVSync48 benchmarks in \Cref{tab:fsl}. Various Ks simulate the scenario with different human curation budgets. Results are averaged over 3 runs with randomly sampled $K$ videos per class for finetuning, except when K equals the number of full training data, in which case, a single run is performed.
For each value of $K$, we treat a "finetuning-only" approach as the baseline paradigm, and show the gains of pre-training as superscripts in the table.
As can be seen, for all values of $K$, pretraining enhances the final generation results, especially with fewer finetuning data. More importantly, our paradigm with minimal human curation efforts (K=1) can achieve stronger performance than baselines trained with 10$\times$ more fine-tuning data. For example, on AVSync15, it performs stronger than K=90 baseline on synchronization (AlignSync: 94.24→94.28) despite higher FVD (277.5→319.7), and can largely surpass K=10 baseline on both FVD (352.6→319.7) and AlignSync (93.32→94.28). On AVSync48, we observe the same trend, where two-stage training with K=1 surpasses K=10 baseline drastically (FVD 116.8→102.5, AlignSync 94.16→94.72). 
These results show that the proposed training paradigm is more budget-efficient thus more scalable than prior works.

\begin{figure}[!t]
    \centering
    \begin{subfigure}[b]{0.49\textwidth}
        \includegraphics[width=\textwidth]{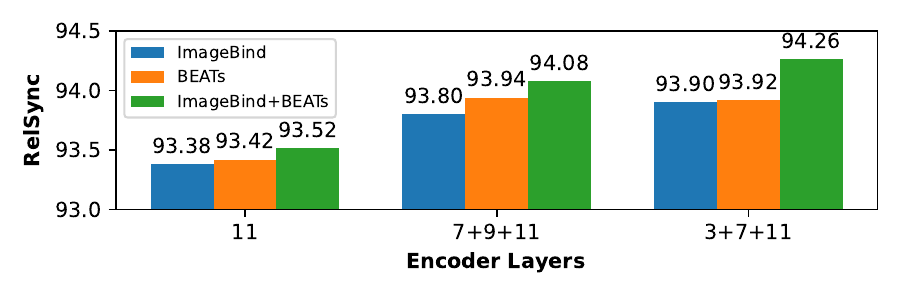}
        \caption{Multi-feature conditioning ablation.}
        \label{fig:ablation-layer}
    \end{subfigure}
    \hfill
    \begin{subfigure}[b]{0.49\textwidth}
        \includegraphics[width=\textwidth]{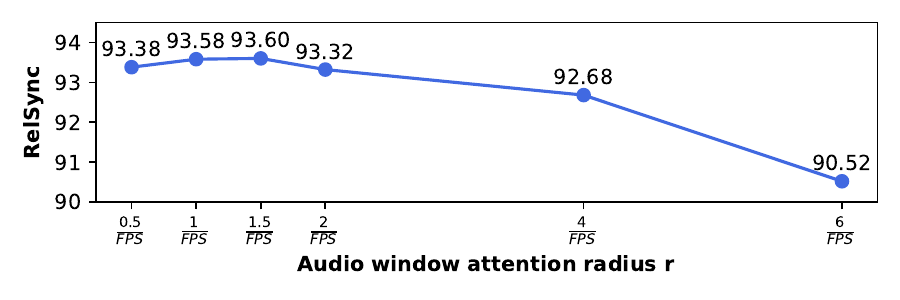}
        \caption{Window attention ablation.}
        \label{fig:ablation-window}
    \end{subfigure}
    \caption{Ablation on proposed audio conditioning modules on AVSync15.}
    \label{fig:ablation_param}
\end{figure}

\paragraph{Audio multi-feature conditioning}
We validate the proposed audio multi-feature conditioning in \Cref{fig:ablation-layer}.
Results show that synchronization performance can be improved by either adopting more layers or using multiple encoders for feature extraction.
Extracting features from layers 3-7-11 demonstrates slightly better performance compared to 7-9-11, suggesting that a more diverse feature granularity is preferred. The optimized audio feature conditioning improves upon the ImageBind-11 baseline as in previous works~\cite{linz2024asva,wang2025keyvid}, significantly improving RelSync from 93.38 to 94.26.

\paragraph{Audio attention window}
As shown in \Cref{fig:ablation-window}, the baseline is a window radius of 0.5/FPS, showing many-to-one correspondence between audio tokens and frames. This heuristic strategy generally adopted in previous works~\cite{linz2024asva,wang2025keyvid} however does not provide the best context window, as larger window radius of 1.5/FPS gives more audio contexts for each frame, increasing RelSync from 93.38 to 93.60. However, further increasing the window leads to contexts that are too broad to focus on local synchronization signals, decreasing the synchronization scores significantly.

\paragraph{Open domain video generation}
\label{sec:ablation open domain}
We demonstrate the strong generalization capability of our model on open-domain ASVA tasks, as shown in \Cref{fig:open-domain}. Results are produced by our 12-FPS model pretraind and then finetuned with K=10 examples per class on AVSync48. Notably, only four of the shown classes are available in AVSync48 (baby crying, lion roaring, fireworks, and explosion), while the rest represent zero-shot settings. 
Given the frozen priors in the pretrained DynamiCrafter and audio encoders, our model can generate highly synchronized video motions following the input audio and text even in zero-shot settings, demonstrating strong controllability on video generation across various domains with natural and accurate physical modeling and visual effects.
For example, in the eating scenario, chewing sound perfectly aligns with the mouth opening/closing action; in the walking scenario, the person alternates arm and leg movement, where the arm and the leg on opposite side move in sync to the footstep sound; in the case of slapping, the person's hand first raises silently, then suddenly drops down with his head turning to the other side, triggering the sound of slapping; in the volcano example, different visual effects can be applied in sync to sounds. We provide more videos in the Appendix, demonstrating the strong potential of our model in open domains.

\section{Conclusion}\label{sec:conclusion}

We proposed an efficient training paradigm to scale up Audio-Synchronized Visual Animation via a two-stage pipeline: large-scale pretraining on automatically curated noisy data, followed by finetuning on a small set of high-quality videos. Our method introduces multi-feature conditioning and window attention for efficient learning and achieves strong results on the manually curated AVSync48 benchmark, which triples the class diversity of prior datasets.

While effective, our model still falls short of simulating real-world video quality, with limitations in resolution, temporal consistency, and motion naturalness. Addressing these requires significantly larger models and datasets, which we leave for future work.

Our approach advances open-domain audio-driven video generation and editing, enabling broader applications beyond talking faces. However, it also raises concerns about misuse, highlighting the need for responsible deployment and safeguards.

\medskip

\small{
\bibliographystyle{style/splncs04}
\bibliography{refs}

\begin{thebibliography}{10}
\providecommand{\url}[1]{\texttt{#1}}
\providecommand{\urlprefix}{URL }
\providecommand{\doi}[1]{https://doi.org/#1}

\bibitem{laionaesthetics}
Laion-aesthetics predictor v2. \url{https://github.com/christophschuhmann/improved-aesthetic-predictor}

\bibitem{quicktime}
Quicktime player. \url{https://support.apple.com/en-us/106375}

\bibitem{llama3modelcard}
AI@Meta: Llama 3 model card  (2024), \url{https://github.com/meta-llama/llama3/blob/main/MODEL_CARD.md}

\bibitem{aytar2016soundnet}
Aytar, Y., Vondrick, C., Torralba, A.: Soundnet: Learning sound representations from unlabeled video. In: Advances in Neural Information Processing Systems (2016)

\bibitem{craft2019}
Baek, Y., Lee, B., Han, D., Yun, S., Lee, H.: Character region awareness for text detection. In: Proceedings of the IEEE Conference on Computer Vision and Pattern Recognition. pp. 9365--9374 (2019)

\bibitem{Bain21webvid10m}
Bain, M., Nagrani, A., Varol, G., Zisserman, A.: Frozen in time: A joint video and image encoder for end-to-end retrieval. In: IEEE International Conference on Computer Vision (2021)

\bibitem{berthelot2019mixmatchholisticapproachsemisupervised}
Berthelot, D., Carlini, N., Goodfellow, I., Papernot, N., Oliver, A., Raffel, C.: Mixmatch: A holistic approach to semi-supervised learning (2019), \url{https://arxiv.org/abs/1905.02249}

\bibitem{pyscenedetect}
Castellano, B.: Pyscenedetect. \url{https://www.scenedetect.com/}

\bibitem{chen2023videocrafter1}
Chen, H., Xia, M., He, Y., Zhang, Y., Cun, X., Yang, S., Xing, J., Liu, Y., Chen, Q., Wang, X., Weng, C., Shan, Y.: Videocrafter1: Open diffusion models for high-quality video generation (2023)

\bibitem{chen2021vggsoundsync}
Chen, H., Xie, W., Afouras, T., Nagrani, A., Vedaldi, A., Zisserman, A.: Audio-visual synchronization in the wild. In: Proceedings of the British Machine Vision Conference (BMVC) (2021)

\bibitem{chen2020vggsound}
Chen, H., Xie, W., Vedaldi, A., Zisserman, A.: Vggsound: A large-scale audio-visual dataset. In: ICASSP (2020)

\bibitem{Chen2022beats}
Chen, S., Wu, Y., Wang, C., Liu, S., Tompkins, D., Chen, Z., Wei, F.: Beats: Audio pre-training with acoustic tokenizers. In: ICML (2023)

\bibitem{chen2020simpleframeworkcontrastivelearning}
Chen, T., Kornblith, S., Norouzi, M., Hinton, G.: A simple framework for contrastive learning of visual representations (2020), \url{https://arxiv.org/abs/2002.05709}

\bibitem{chen2024panda70m}
Chen, T.S., Siarohin, A., Menapace, W., Deyneka, E., Chao, H.w., Jeon, B.E., Fang, Y., Lee, H.Y., Ren, J., Yang, M.H., Tulyakov, S.: Panda-70m: Captioning 70m videos with multiple cross-modality teachers. In: Proceedings of the IEEE/CVF Conference on Computer Vision and Pattern Recognition (2024)

\bibitem{chen2020mocov2}
Chen, X., Fan, H., Girshick, R., He, K.: Improved baselines with momentum contrastive learning. arXiv preprint arXiv:2003.04297  (2020)

\bibitem{vicuna2023}
Chiang, W.L., Li, Z., Lin, Z., Sheng, Y., Wu, Z., Zhang, H., Zheng, L., Zhuang, S., Zhuang, Y., Gonzalez, J.E., Stoica, I., Xing, E.P.: Vicuna: An open-source chatbot impressing gpt-4 with 90\%* chatgpt quality (March 2023), \url{https://lmsys.org/blog/2023-03-30-vicuna/}

\bibitem{lipreading}
Chung, J.S., Zisserman, A.: Lip reading in the wild. In: ACCV (2016)

\bibitem{outoftime}
Chung, J.S., Zisserman, A.: Out of time: automated lip sync in the wild. In: ACCV Workshop (2016)

\bibitem{Chung2019perfectmatch}
Chung, S.W., Chung, J.S., Kang, H.G.: Perfect match: Improved cross-modal embeddings for audio-visual synchronisation. In: ICASSP 2019 - 2019 IEEE International Conference on Acoustics, Speech and Signal Processing (ICASSP) (2019)

\bibitem{crowson2024sd3}
Crowson, K., Baumann, S.A., Birch, A., Abraham, T.M., Kaplan, D.Z., Shippole, E.: Scalable high-resolution pixel-space image synthesis with hourglass diffusion transformers (2024), \url{https://arxiv.org/abs/2401.11605}

\bibitem{ebenezer2021detectionavsync}
Ebenezer, J.P., Wu, Y., Wei, H., Sethuraman, S., Liu, Z.: Detection of audio-video synchronization errors via event detection. In: ICASSP (2021)

\bibitem{audioset}
Gemmeke, J.F., Ellis, D.P.W., Freedman, D., Jansen, A., Lawrence, W., Moore, R.C., Plakal, M., Ritter, M.: Audio set: An ontology and human-labeled dataset for audio events. In: Proc. IEEE ICASSP 2017 (2017)

\bibitem{girdhar2023imagebind}
Girdhar, R., El-Nouby, A., Liu, Z., Singh, M., Alwala, K.V., Joulin, A., Misra, I.: Imagebind: One embedding space to bind them all. In: CVPR (2023)

\bibitem{NIPS2004entropymin}
Grandvalet, Y., Bengio, Y.: Semi-supervised learning by entropy minimization. In: Saul, L., Weiss, Y., Bottou, L. (eds.) Advances in Neural Information Processing Systems. vol.~17. MIT Press (2004), \url{https://proceedings.neurips.cc/paper_files/paper/2004/file/96f2b50b5d3613adf9c27049b2a888c7-Paper.pdf}

\bibitem{HaCohen2024LTXVideo}
HaCohen, Y., Chiprut, N., Brazowski, B., Shalem, D., Moshe, D., Richardson, E., Levin, E., Shiran, G., Zabari, N., Gordon, O., Panet, P., Weissbuch, S., Kulikov, V., Bitterman, Y., Melumian, Z., Bibi, O.: Ltx-video: Realtime video latent diffusion. arXiv preprint arXiv:2501.00103  (2024)

\bibitem{MaskedAutoencoders2021}
He, K., Chen, X., Xie, S., Li, Y., Doll{\'a}r, P., Girshick, R.: Masked autoencoders are scalable vision learners. arXiv:2111.06377  (2021)

\bibitem{he2021maskedautoencodersscalablevision}
He, K., Chen, X., Xie, S., Li, Y., Dollár, P., Girshick, R.: Masked autoencoders are scalable vision learners (2021), \url{https://arxiv.org/abs/2111.06377}

\bibitem{he2019moco}
He, K., Fan, H., Wu, Y., Xie, S., Girshick, R.: Momentum contrast for unsupervised visual representation learning. arXiv preprint arXiv:1911.05722  (2019)

\bibitem{ho2020ddpm}
Ho, J., Jain, A., Abbeel, P.: Denoising diffusion probabilistic models. In: NeurIPS (2020)

\bibitem{hu2021loralowrankadaptationlarge}
Hu, E.J., Shen, Y., Wallis, P., Allen-Zhu, Z., Li, Y., Wang, S., Wang, L., Chen, W.: Lora: Low-rank adaptation of large language models (2021), \url{https://arxiv.org/abs/2106.09685}

\bibitem{huang2022amae}
Huang, P.Y., Xu, H., Li, J., Baevski, A., Auli, M., Galuba, W., Metze, F., Feichtenhofer, C.: Masked autoencoders that listen. In: NeurIPS (2022)

\bibitem{sparse2022iashin}
Iashin, V., Xie, W., Rahtu, E., Zisserman, A.: Sparse in space and time: Audio-visual synchronisation with trainable selectors. In: British Machine Vision Conference (BMVC) (2022)

\bibitem{jeong2023tpos}
Jeong, Y., Ryoo, W., Lee, S., Seo, D., Byeon, W., Kim, S., Kim, J.: The power of sound (tpos): Audio reactive video generation with stable diffusion. In: Proceedings of the IEEE/CVF International Conference on Computer Vision. pp. 7822--7832 (2023)

\bibitem{text2video-zero}
Khachatryan, L., Movsisyan, A., Tadevosyan, V., Henschel, R., Wang, Z., Navasardyan, S., Shi, H.: Text2video-zero: Text-to-image diffusion models are zero-shot video generators. In: ICCV (2023)

\bibitem{audiocaps}
Kim, C.D., Kim, B., Lee, H., Kim, G.: Audiocaps: Generating captions for audios in the wild. In: NAACL-HLT (2019)

\bibitem{kingma2017adamoptimizer}
Kingma, D.P., Ba, J.: Adam: A method for stochastic optimization (2017), \url{https://arxiv.org/abs/1412.6980}

\bibitem{Koch2015SiameseNN}
Koch, G.R.: Siamese neural networks for one-shot image recognition (2015), \url{https://api.semanticscholar.org/CorpusID:13874643}

\bibitem{kong2024hunyuanvideo}
Kong, W., Tian, Q., Zhang, Z., Min, R., Dai, Z., Zhou, J., Xiong, J., Li, X., Wu, B., Zhang, J., et~al.: Hunyuanvideo: A systematic framework for large video generative models. arXiv preprint arXiv:2412.03603  (2024)

\bibitem{NEURIPS2018coopav}
Korbar, B., Tran, D., Torresani, L.: Cooperative learning of audio and video models from self-supervised synchronization. In: Advances in Neural Information Processing Systems (2018)

\bibitem{laine2017temporalensemblingsemisupervisedlearning}
Laine, S., Aila, T.: Temporal ensembling for semi-supervised learning (2017), \url{https://arxiv.org/abs/1610.02242}

\bibitem{lee2022soundguided}
Lee, S.H., Oh, G., Byeon, W., Bae, J., Kim, C., Ryoo, W.J., Yoon, S.H., Kim, J., Kim, S.: Sound-guided semantic video generation. In: ECCV (2022)

\bibitem{lee2023aadiff}
Lee, S., Kong, C., Jeon, D., Kwak, N.: Aadiff: Audio-aligned video synthesis with text-to-image diffusion. In: CVPR Workshop on Content Generation (2023)

\bibitem{active1}
Lewis, D.D., Gale, W.A.: A sequential algorithm for training text classifiers. In: Proceedings of the 17th Annual International ACM SIGIR Conference on Research and Development in Information Retrieval. p. 3–12. SIGIR '94, Springer-Verlag, Berlin, Heidelberg (1994)

\bibitem{li2021aist++}
Li, R., Yang, S., Ross, D.A., Kanazawa, A.: Learn to dance with aist++: Music conditioned 3d dance generation. In: ICCV (2021)

\bibitem{li2023gligen}
Li, Y., Liu, H., Wu, Q., Mu, F., Yang, J., Gao, J., Li, C., Lee, Y.J.: Gligen: Open-set grounded text-to-image generation. In: CVPR (2023)

\bibitem{Liu_2024_CVPR_llava15}
Liu, H., Li, C., Li, Y., Lee, Y.J.: Improved baselines with visual instruction tuning. In: Proceedings of the IEEE/CVF Conference on Computer Vision and Pattern Recognition (CVPR). pp. 26296--26306 (June 2024)

\bibitem{liu2023llava}
Liu, H., Li, C., Wu, Q., Lee, Y.J.: Visual instruction tuning. In: NeurIPS (2023)

\bibitem{mei2023wavcaps}
Mei, X., Meng, C., Liu, H., Kong, Q., Ko, T., Zhao, C., Plumbley, M.D., Zou, Y., Wang, W.: Wav{C}aps: A {ChatGPT}-assisted weakly-labelled audio captioning dataset for audio-language multimodal research. IEEE/ACM Transactions on Audio, Speech, and Language Processing pp. 1--15 (2024)

\bibitem{mou2023t2iadapterlearningadaptersdig}
Mou, C., Wang, X., Xie, L., Wu, Y., Zhang, J., Qi, Z., Shan, Y., Qie, X.: T2i-adapter: Learning adapters to dig out more controllable ability for text-to-image diffusion models (2023), \url{https://arxiv.org/abs/2302.08453}

\bibitem{ng2024audio2photoreal}
Ng, E., Romero, J., Bagautdinov, T., Bai, S., Darrell, T., Kanazawa, A., Richard, A.: From audio to photoreal embodiment: Synthesizing humans in conversations. In: ArXiv (2024)

\bibitem{owens2016greatesthits}
Owens, A., Isola, P., McDermott, J., Torralba, A., Adelson, E.H., Freeman, W.T.: Visually indicated sounds. In: CVPR (2016)

\bibitem{park2022synctalkface}
Park, S.J., Kim, M., Hong, J., Choi, J., Ro, Y.M.: Synctalkface: Talking face generation with precise lip-syncing via audio-lip memory. In: AAAI Conference on Artificial Intelligence (AAAI) (2022)

\bibitem{polyak2025moviegen}
Polyak, A., Zohar, A., Brown, A., Tjandra, A., Sinha, A., Lee, A., Vyas, A., Shi, B., Ma, C.Y., Chuang, C.Y., Yan, D., Choudhary, D., Wang, D., Sethi, G., Pang, G., Ma, H., Misra, I., Hou, J., Wang, J., Jagadeesh, K., Li, K., Zhang, L., Singh, M., Williamson, M., Le, M., Yu, M., Singh, M.K., Zhang, P., Vajda, P., Duval, Q., Girdhar, R., Sumbaly, R., Rambhatla, S.S., Tsai, S., Azadi, S., Datta, S., Chen, S., Bell, S., Ramaswamy, S., Sheynin, S., Bhattacharya, S., Motwani, S., Xu, T., Li, T., Hou, T., Hsu, W.N., Yin, X., Dai, X., Taigman, Y., Luo, Y., Liu, Y.C., Wu, Y.C., Zhao, Y., Kirstain, Y., He, Z., He, Z., Pumarola, A., Thabet, A., Sanakoyeu, A., Mallya, A., Guo, B., Araya, B., Kerr, B., Wood, C., Liu, C., Peng, C., Vengertsev, D., Schonfeld, E., Blanchard, E., Juefei-Xu, F., Nord, F., Liang, J., Hoffman, J., Kohler, J., Fire, K., Sivakumar, K., Chen, L., Yu, L., Gao, L., Georgopoulos, M., Moritz, R., Sampson, S.K., Li, S., Parmeggiani, S., Fine, S., Fowler, T., Petrovic, V., Du, Y.: Movie gen: A cast of
  media foundation models (2025), \url{https://arxiv.org/abs/2410.13720}

\bibitem{radford2021CLIP}
Radford, A., Kim, J.W., Hallacy, C., Ramesh, A., Goh, G., Agarwal, S., Sastry, G., Askell, A., Mishkin, P., Clark, J., Krueger, G., Sutskever, I.: Learning transferable visual models from natural language supervision. In: ICML (2021)

\bibitem{robin2022ldm}
Rombach, R., Blattmann, A., Lorenz, D., Esser, P., Ommer, B.: High-resolution image synthesis with latent diffusion models. In: CVPR (2022)

\bibitem{sener2018activelearningconvolutionalneural}
Sener, O., Savarese, S.: Active learning for convolutional neural networks: A core-set approach (2018), \url{https://arxiv.org/abs/1708.00489}

\bibitem{active2}
Seung, H.S., Opper, M., Sompolinsky, H.: Query by committee. In: Proceedings of the Fifth Annual Workshop on Computational Learning Theory. p. 287–294. COLT '92, Association for Computing Machinery, New York, NY, USA (1992). \doi{10.1145/130385.130417}, \url{https://doi.org/10.1145/130385.130417}

\bibitem{singer2022makeavideo}
Singer, U., Polyak, A., Hayes, T., Yin, X., An, J., Zhang, S., Hu, Q., Yang, H., Ashual, O., Gafni, O., Parikh, D., Gupta, S., Taigman, Y.: Make-a-video: Text-to-video generation without text-video data (2022)

\bibitem{snell2017prototypicalnetworksfewshotlearning}
Snell, J., Swersky, K., Zemel, R.S.: Prototypical networks for few-shot learning (2017), \url{https://arxiv.org/abs/1703.05175}

\bibitem{sohn2020fixmatchsimplifyingsemisupervisedlearning}
Sohn, K., Berthelot, D., Li, C.L., Zhang, Z., Carlini, N., Cubuk, E.D., Kurakin, A., Zhang, H., Raffel, C.: Fixmatch: Simplifying semi-supervised learning with consistency and confidence (2020), \url{https://arxiv.org/abs/2001.07685}

\bibitem{song2022ddim}
Song, J., Meng, C., Ermon, S.: Denoising diffusion implicit models. In: ICLR (2021)

\bibitem{sung2018learningcomparerelationnetwork}
Sung, F., Yang, Y., Zhang, L., Xiang, T., Torr, P.H.S., Hospedales, T.M.: Learning to compare: Relation network for few-shot learning (2018), \url{https://arxiv.org/abs/1711.06025}

\bibitem{tan2024ominicontrol}
Tan, Z., Liu, S., Yang, X., Xue, Q., Wang, X.: Ominicontrol: Minimal and universal control for diffusion transformer. arXiv preprint arXiv:2411.15098  (2024)

\bibitem{tan2025ominicontrol2}
Tan, Z., Xue, Q., Yang, X., Liu, S., Wang, X.: Ominicontrol2: Efficient conditioning for diffusion transformers. arXiv preprint arXiv:2503.08280  (2025)

\bibitem{tang2023codi}
Tang, Z., Yang, Z., Zhu, C., Zeng, M., Bansal, M.: Any-to-any generation via composable diffusion. In: Thirty-seventh Conference on Neural Information Processing Systems (2023), \url{https://openreview.net/forum?id=2EDqbSCnmF}

\bibitem{tarvainen2018meanteachersbetterrole}
Tarvainen, A., Valpola, H.: Mean teachers are better role models: Weight-averaged consistency targets improve semi-supervised deep learning results (2018), \url{https://arxiv.org/abs/1703.01780}

\bibitem{teed2020raftrecurrentallpairsfield}
Teed, Z., Deng, J.: Raft: Recurrent all-pairs field transforms for optical flow (2020), \url{https://arxiv.org/abs/2003.12039}

\bibitem{aist-dance-db}
Tsuchida, S., Fukayama, S., Hamasaki, M., Goto, M.: Aist dance video database: Multi-genre, multi-dancer, and multi-camera database for dance information processing. In: Proceedings of the 20th International Society for Music Information Retrieval Conference, {ISMIR} 2019. Delft, Netherlands (Nov 2019)

\bibitem{fvd}
Unterthiner, T., van Steenkiste, S., Kurach, K., Marinier, R., Michalski, M., Gelly, S.: Towards accurate generative models of video: A new metric \& challenges. In: arXiv (2019)

\bibitem{vaswani2023attention}
Vaswani, A., Shazeer, N., Parmar, N., Uszkoreit, J., Jones, L., Gomez, A.N., Kaiser, L., Polosukhin, I.: Attention is all you need. In: NeurIPS (2017)

\bibitem{vinyals2017matchingnetworksshotlearning}
Vinyals, O., Blundell, C., Lillicrap, T., Kavukcuoglu, K., Wierstra, D.: Matching networks for one shot learning (2017), \url{https://arxiv.org/abs/1606.04080}

\bibitem{wan2025}
Wang, A., Ai, B., Wen, B., Mao, C., Xie, C.W., Chen, D., Yu, F., Zhao, H., Yang, J., Zeng, J., Wang, J., Zhang, J., Zhou, J., Wang, J., Chen, J., Zhu, K., Zhao, K., Yan, K., Huang, L., Feng, M., Zhang, N., Li, P., Wu, P., Chu, R., Feng, R., Zhang, S., Sun, S., Fang, T., Wang, T., Gui, T., Weng, T., Shen, T., Lin, W., Wang, W., Wang, W., Zhou, W., Wang, W., Shen, W., Yu, W., Shi, X., Huang, X., Xu, X., Kou, Y., Lv, Y., Li, Y., Liu, Y., Wang, Y., Zhang, Y., Huang, Y., Li, Y., Wu, Y., Liu, Y., Pan, Y., Zheng, Y., Hong, Y., Shi, Y., Feng, Y., Jiang, Z., Han, Z., Wu, Z.F., Liu, Z.: Wan: Open and advanced large-scale video generative models. arXiv preprint arXiv:2503.20314  (2025)

\bibitem{wang2023modelscope}
Wang, J., Yuan, H., Chen, D., Zhang, Y., Wang, X., Zhang, S.: Modelscope text-to-video technical report (2023)

\bibitem{vid2vid-zero}
Wang, W., Xie, k., Liu, Z., Chen, H., Cao, Y., Wang, X., Shen, C.: Zero-shot video editing using off-the-shelf image diffusion models. arXiv preprint arXiv:2303.17599  (2023)

\bibitem{wang2025keyvid}
Wang, X., Liu, J., Wang, Z., Yu, X., Wu, J., Sun, X., Su, Y., Yuille, A., Liu, Z., Barsoum, E.: Keyvid: Keyframe-aware video diffusion for audio-synchronized visual animation (2025), \url{https://arxiv.org/abs/2504.09656}

\bibitem{wang2024internvidlargescalevideotextdataset}
Wang, Y., He, Y., Li, Y., Li, K., Yu, J., Ma, X., Li, X., Chen, G., Chen, X., Wang, Y., He, C., Luo, P., Liu, Z., Wang, Y., Wang, L., Qiao, Y.: Internvid: A large-scale video-text dataset for multimodal understanding and generation (2024), \url{https://arxiv.org/abs/2307.06942}

\bibitem{wu2023lamp}
Wu, R., Chen, L., Yang, T., Guo, C., Li, C., Zhang, X.: Lamp: Learn a motion pattern by few-shot tuning a text-to-image diffusion model. arXiv preprint arXiv:2310.10769  (2023)

\bibitem{xie2022simmimsimpleframeworkmasked}
Xie, Z., Zhang, Z., Cao, Y., Lin, Y., Bao, J., Yao, Z., Dai, Q., Hu, H.: Simmim: A simple framework for masked image modeling (2022), \url{https://arxiv.org/abs/2111.09886}

\bibitem{xing2023dynamicrafter}
Xing, J., Xia, M., Zhang, Y., Chen, H., Yu, W., Liu, H., Wang, X., Wong, T.T., Shan, Y.: Dynamicrafter: Animating open-domain images with video diffusion priors. arXiv preprint arXiv:2310.12190  (2023)

\bibitem{xue2022hdvila}
Xue, H., Hang, T., Zeng, Y., Sun, Y., Liu, B., Yang, H., Fu, J., Guo, B.: Advancing high-resolution video-language representation with large-scale video transcriptions. In: International Conference on Computer Vision and Pattern Recognition (CVPR) (2022)

\bibitem{yariv2023tempotoken}
Yariv, G., Gat, I., Benaim, S., Wolf, L., Schwartz, I., Adi, Y.: Diverse and aligned audio-to-video generation via text-to-video model adaptation (2023)

\bibitem{ye2023geneface}
Ye, Z., Jiang, Z., Ren, Y., Liu, J., He, J., Zhao, Z.: Geneface: Generalized and high-fidelity audio-driven 3d talking face synthesis. In: ICLR (2023)

\bibitem{linz2024asva}
Zhang, L., Mo, S., Zhang, Y., Morgado, P.: Audio-synchronized visual animation. In: Proceedings of the European Conference on Computer Vision (ECCV) (2024)

\bibitem{zhang2023adding}
Zhang, L., Rao, A., Agrawala, M.: Adding conditional control to text-to-image diffusion models. In: IEEE International Conference on Computer Vision (ICCV) (2023)

\bibitem{zhao2023motiondirector}
Zhao, R., Gu, Y., Wu, J.Z., Zhang, D.J., Liu, J., Wu, W., Keppo, J., Shou, M.Z.: Motiondirector: Motion customization of text-to-video diffusion models. arXiv preprint arXiv:2310.08465  (2023)

\bibitem{zhou2019talking}
Zhou, H., Liu, Y., Liu, Z., Luo, P., Wang, X.: Talking face generation by adversarially disentangled audio-visual representation. In: AAAI Conference on Artificial Intelligence (AAAI) (2019)

\bibitem{Zhou2021Pose}
Zhou, H., Sun, Y., Wu, W., Loy, C.C., Wang, X., Liu, Z.: Pose-controllable talking face generation by implicitly modularized audio-visual representation. In: Proceedings of the IEEE Conference on Computer Vision and Pattern Recognition (CVPR) (2021)

\bibitem{zhu2023minigpt4enhancingvisionlanguageunderstanding}
Zhu, D., Chen, J., Shen, X., Li, X., Elhoseiny, M.: Minigpt-4: Enhancing vision-language understanding with advanced large language models (2023), \url{https://arxiv.org/abs/2304.10592}

\end{thebibliography}
}




\end{document}